# Optimizing YOLOv5s Object Detection through Knowledge Distillation algorithm

Guanming Huang
University of Chicago
Chicago, USA

Aoran Shen
University of Michigan
Ann Arbor, USA

Yuxiang Hu
Johns Hopkins University
Baltimore，USA

Junliang Du
Shanghai Jiao Tong University
Shanghai, China

Jiacheng Hu
Tulane University
New Orleans, USA

Yingbin Liang*
Northeastern University
Seattle, USA

*Abstract*— **This paper explores the application of knowledge distillation technology in target detection tasks, especially the impact of different distillation temperatures on the performance of student models. By using YOLOv5l as the teacher network and a smaller YOLOv5s as the student network, we found that with the increase of distillation temperature, the student's detection accuracy gradually improved, and finally achieved mAP50 and mAP50-95 indicators that were better than the original YOLOv5s model at a specific temperature. Experimental results show that appropriate knowledge distillation strategies can not only improve the accuracy of the model but also help improve the reliability and stability of the model in practical applications. This paper also records in detail the accuracy curve and loss function descent curve during the model training process and shows that the model converges to a stable state after 150 training cycles. These findings provide a theoretical basis and technical reference for further optimizing target detection algorithms.**

## I. INTRODUCTION

The rapid advancement of deep learning, big data, and hardware technology has led computers to increasingly replace traditional manual processes, making them essential tools for information acquisition. Artificial intelligence (AI), widely recognized as a driver of industrial innovation, has been central to this transformation. Within AI, computer vision [1] has become a critical field, receiving significant attention from researchers due to its potential to revolutionize how machines interpret visual data.

A fundamental task in computer vision is object detection, which has driven the advancement of related technologies. Historically, object detection has evolved through two key phases: classical algorithms and modern deep learning-based techniques [2]. These developments not only advance scientific research but also have practical value across various fields. For example, object detection plays a crucial role in medical imaging [3], text extraction [4-6], and named entity recognition [7], demonstrating its applicability in specialized areas.

The advancements achieved by YOLOv5s exemplify the broader progress made in object detection technology [8], highlighting its importance in numerous industries. The ongoing development of detection algorithms like YOLOv5s has been complemented by the creation of comprehensive, annotated datasets that are essential for training AI models. These datasets, combined with YOLOv5s' efficiency in handling large-scale detection tasks, have allowed for more accurate and reliable object detection systems. This has not only improved performance in real-time applications but also enhanced safety and operational efficiency in fields such as autonomous driving, industrial automation, and healthcare [9]. This article studies deep learning networks and constructs a localization technique and an object detection algorithm that combines the extraction of location information of the objects to be detected, in order to meet the application requirements in different scenarios. The main work of this article includes:

(1) A detection algorithm combining location information and knowledge extraction is proposed to solve the problem of location blur in object detection in image tasks.

(2) The bounding box is converted into a probability distribution so that the teacher model provides the student model with richer localization knowledge. A region-based selective distillation strategy is proposed to select the distillation area on the feature map based on the position information.

(3) Extract classification knowledge and localization knowledge separately from the feature map, and decouple the feature map into classification and localization heads to further improve the performance of object detection.

## II. RELATED WORK

Recent advances in deep learning have significantly influenced object detection algorithms, particularly through the development of optimization strategies and knowledge transfer mechanisms. The use of knowledge distillation [10-12], a method of transferring information from a larger, well-trained model to a smaller one, has emerged as a promising technique for improving model performance while maintaining computational efficiency [13]. Research into optimization strategies has addressed common challenges such as reducing training bias and enhancing the convergence speed of deep learning models, both of which are crucial for the effective application of knowledge distillation in object detection.

A key contribution to deep learning optimization is the focus on reducing bias in the training process [14]. Such advancements have led to more reliable and robust models, which are essential when applying knowledge distillation to smaller networks like YOLOv5s. Techniques that enhance gradient-based optimization significantly improve the stability and performance of student models, ensuring more effective knowledge transfer from the teacher model. This directly impacts the efficiency of the distillation process, which is highly sensitive to optimization parameters such as learning rates and momentum.

Moreover, deep learning methods for feature extraction have advanced, particularly in the domain of convolutional neural networks (CNNs), which are fundamental to object detection tasks [15]. Effective feature extraction mechanisms enable student models to capture more detailed localization and classification knowledge from the teacher model. Recent developments in feature extraction techniques, especially those that focus on improving spatial resolution, enhance the accuracy of bounding boxes and object localization—a core challenge in object detection. Yan et al. [16] explore image super-resolution techniques based on convolutional neural networks (CNNs), which further underline the importance of fine-tuning deep learning models for enhanced feature extraction, analogous to the extraction of localization knowledge in object detection models.

The decoupling of feature extraction into classification and localization tasks, as explored in object detection models, aligns with innovations in structured knowledge distillation [17]. By isolating these two aspects, the student model can gain a more refined understanding of object placement, which is especially beneficial in real-time detection tasks. The introduction of selective distillation strategies further optimizes this process by focusing on the most informative areas of the feature map, reducing computational overhead while maximizing performance gains.

Another significant impact of recent work in deep learning involves the application of contrastive learning and hybrid models [18], which enhance feature representations and improve generalization in small models [19]. These advancements have contributed to more robust training processes, allowing student models to achieve higher accuracy with fewer resources [20]. The implementation of such methods in knowledge distillation frameworks ensures that the student model retains essential features from the teacher model while remaining computationally efficient.

### III.  METHOD

In this section, the proposed target detection algorithm combined with position information distillation will be introduced. This method first selects the key distillation areas and expandable position areas on the multi-layer feature maps of YOLOv5l and YOLOv5s, and then decouples the multi-layer feature maps of the two models into the classification detection head and the positioning detection head respectively, and then extracts the classification knowledge and positioning knowledge respectively. For classification knowledge, knowledge distillation is performed on the classification head [21]; for positioning knowledge, the knowledge transfer process of the feature map positioning head is re-described, and the bounding box is switched to the probability distribution. The extraction of these two types of knowledge is based on the logic of a single head, rather than deep features. The overall framework of the network is shown in Figure 1.

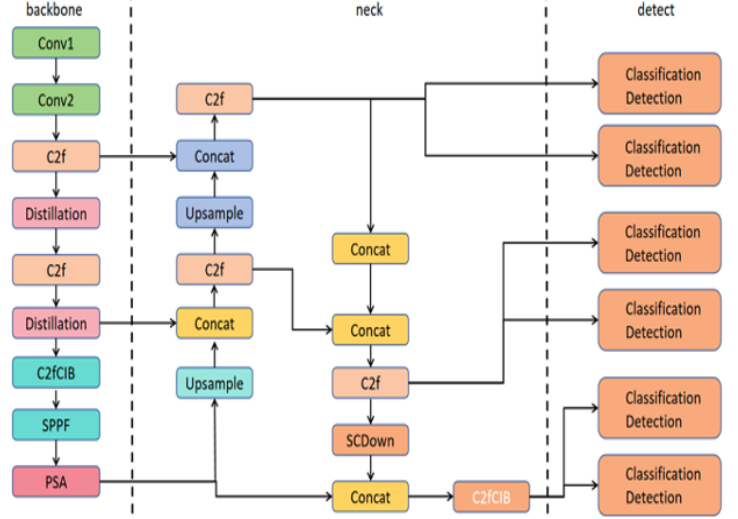

Figure 1 Overall framework of the network

The proposed model processes the input image through the backbone network, extracting multi-scale features. The Feature Pyramid Network (FPN) then handles multi-layer feature maps from the YOLOv5l and YOLOv5s models. During this, channels increase while feature map sizes decrease. Key Distillation Regions (KDR) and Expandable Location Regions (ELR) are identified. The feature map is then split into a classification and localization detection head, followed by top-down and bottom-up fusion at the output.

When processing the feature layers, different-sized anchors are configured to predict targets of varying dimensions. Following this, the output stage generates bounding boxes, category probabilities, and center points to define the position and class of the targets within the image. Both the teacher and student models apply non-maximum suppression to eliminate redundant or overlapping bounding boxes, retaining only the most reliable target predictions. The teacher model undergoes training for knowledge distillation to acquire optimal training weights. Backpropagation is then utilized to refine the student network model, optimizing it through the feedback loop. This process culminates in the calculation of an overall distillation loss function, which encapsulates the refinement of the student model through the guidance provided by the teacher model.

The representation of bounding boxes has undergone a transformative journey, evolving from a simplistic Dirac-delta distribution to a more nuanced Gaussian distribution, and further still to a sophisticated probability distribution. This progression reflects a deeper understanding of the variability in object shapes and positions. With the probability distribution approach, the system can now predict a range of possibilities, including single-peak distributions, bimodal distributions, or even multi-peak distributions. This advanced representation allows for the encoding of both ambiguous and well-defined boundaries of objects. The flatness or sharpness of the distribution can effectively convey the uncertainty or certainty of the object's edges. Consequently, position information distillation benefits from this probabilistic approach, as it can now represent both blurry and sharply defined edges, thereby providing a richer and more accurate description of object locations within an image.

For a given bounding box B, there are two traditional representations, namely {x, y, h, w} and {t, b, l, r}. However, both of these forms only focus on the position of the true value and cannot model the ambiguity of the bounding box.

Use B = {t, B, l, r} to represent the bounding box. Let $e \in B$ be an edge of the bounding box. Its value can usually be expressed as:

$$e = \int_{e_{\min}}^{e_{\max}} x \Pr(x) dx, \quad e \in B$$

Among them, the range of regression coordinates is $[e_{\min}, e_{\max}]$, and $\Pr(x)$ is the corresponding probability. In the above formula, when $x = e^{gt}$, $\Pr(x) = 1$, otherwise $\Pr(x) = 0$. By quantizing the continuous regression range $[e_{\min}, e_{\max}]$ into a uniform discrete variable $[e_1, e_2, ..., e_n]$, the formula has n subintervals, where $e_1 = e_{\min}$, $e_n = e_{\max}$ and each given edge are $\Pr(x)$, by using the softmax function to represent each edge of a given bounding box as a probability distribution. Therefore, the probability distribution of the 4 bounding boxes in the bounding box can be used to measure the instability of the prediction. The overall schematic diagram is shown in the figure below:

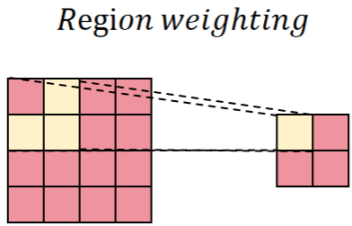

Figure 2 Region weighting diagram

We give the relevant loss function for classification. The loss function is as follows:

$$L_{cls}^{md} = \gamma L_{cls}^{CE}(Ps, Pt) + (1-\gamma) L_{cls}^{CE}(Ps, y_{cls})$$

$$L_{cls}^{CE}(Ps, Pt) = -\sum Pt * \log(Ps)$$

$$L_{cls}^{CE}(Ps, y_{cls}) = -\sum y_{cls} * \log(Ps)$$

The first formula is the loss function of classification prediction in the key distillation area, the second formula is the cross entropy loss between the classification prediction of the network we proposed and the classification prediction of the distillation network, and the third formula is the cross entropy loss between the classification result predicted by the student network and the true label.

Finally, our confidence loss function formula is as follows:

$$L_{obj} = L_{obj}(O_s, y_{obj}) + \varepsilon L_{obj}(O_s, O_t)$$

$$L_{obj}(O_s, y_{obj}) = \frac{1}{n} \sum (O_s - y_{obj})^2$$

$$L_{obj}(O_s, O_t) = \frac{1}{n} \sum (O_s - O_t)^2$$

IV. EXPERIMENT

A. Experimental setup

The hardware configuration information used in the experiment is shown in Table 1.

Table 1 Experimental hardware parameters

| Experimental environment | Related parameters |
|---|---|
| operating system | Ubuntu22.04 |
| CPU | Inter(R) Core(TM) i9-14700kf |
| Memory size | 64G |
| GPU | NVIDIA GeForce RTX 3090 |
| programming language | Python3.11 |
| Deep Learning Frameworks | Pytorch 2.0 |

The experimental training parameters are set as follows: During the warmup period, the learning rate is updated by one-dimensional linear interpolation. After the warmup phase, the subsequent learning rate update adopts the cosine annealing algorithm, and the training of the entire experimental model will last for 300 epochs.

B. Datasets

The experiments in this chapter selected the public COCO dataset, which contains 200,000 images and 80 categories, including over 500,000 object labels. It is widely regarded as the most recognized database for object detection. In this study, 6,000 images were randomly selected and divided into training, validation, and test sets in an 8:1:1 ratio. To evaluate the effectiveness of the improvements made to the original YOLOv5 network model, the experiments employed three key metrics: precision (P), mean average precision (mAP) across categories, and recall (R). An example of the dataset is provided in Figure 3.

Additionally, the linked data techniques referenced in this study played a crucial role in optimizing the dataset's organization and accessibility [22]. By leveraging these techniques, the method improved the integration of diverse data sources, enabling more efficient categorization and annotation of the COCO dataset. This structured approach not only facilitated seamless data processing but also ensured that key object detection features were well-organized and accessible for comprehensive analysis. In particular, the linked data methodology supported the effective partitioning of the 6,000 images into the respective sets, providing a robust framework for assessing the YOLOv5 model enhancements. This systematic approach simplified dataset management and enhanced the application of the precision (P), mAP, and recall (R) metrics, contributing to more accurate and scalable validation of the model's performance. An example of the dataset is shown in Figure 3.

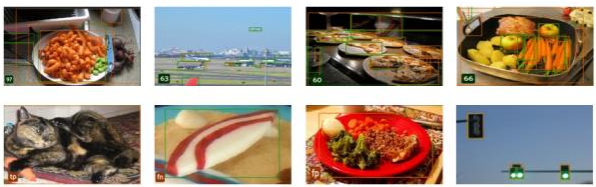

Figure 3 Coco dataset example

*C. Experimental Results*

To investigate the impact of varying distillation temperatures on the detection performance of the student model, this chapter employs YOLOv5l as the teacher network and YOLOv5s as the student network. Initially, the teacher model undergoes training to achieve a high level of proficiency. Subsequently, knowledge distillation technology is utilized to transfer the learned knowledge from the teacher to the student model. During the distillation process, soft labels and attention maps are the primary tools for transferring knowledge. Soft labels provide more detailed location and category information compared to hard labels, aiding the student model in capturing a broader spectrum of data nuances. Attention maps assist the student network in focusing on the most relevant features of the target, thereby enhancing its ability to learn and replicate the teacher's proficiency.

To evaluate the effectiveness of different distillation temperatures, experiments are conducted by adjusting the temperature parameter. These experiments compare the performance of the student model trained with varying temperatures against the original baseline model, the YOLOv5s network. The experimental outcomes, detailing the impact of different distillation temperatures on the detection accuracy and overall performance, are summarized in Table 2. This systematic approach allows for a thorough analysis of how temperature tuning influences the student model's ability to absorb and apply the knowledge distilled from the teacher model.

Table 2 Experimental Results

| Model | Distillation temperature | mAP50 | Map50-95 |
|---|---|---|---|
| YOLOV5s | - | 91.33 | 67.86 |
| Ours-1 | 25 | 93.21 | 70.21 |
| Ours-2 | 30 | 94.56 | 71.77 |
| Ours-3 | 35 | 95.13 | 72.89 |
| Ours-4 | 40 | 96.44 | 73.97 |
| Ours-5 | 45 | 96.75 | 74.56 |

According to the experimental results, it can be seen that at different distillation temperatures, our models (Ours-1 to Ours-5) show significant performance improvements compared to the YOLOV5s model. Specifically, as the distillation temperature increases from 25 to 45 degrees, mAP50 gradually increases from 93.21% to 96.75%, and mAP50-95 also increases from 70.21% to 74.56%, which shows that adjusting the distillation temperature can effectively improve the model. Recognition accuracy. Especially on the high-threshold mAP50-95 index, our model shows stronger generalization ability and more stable detection effect, which is up to about 6.7 percentage points higher than YOLOV5s' 67.86%. This trend not only verifies the effectiveness of the distillation strategy, but also proves the importance of temperature as a hyperparameter. Overall, the experimental results show that by finely adjusting the distillation temperature, the average accuracy of the model under different IoU thresholds can be further enhanced while maintaining a high mAP50, thereby making the model more suitable for complex scenarios in practical applications and providing A more reliable target detection solution.

Furthermore, we give the relevant curves during the training experiment.

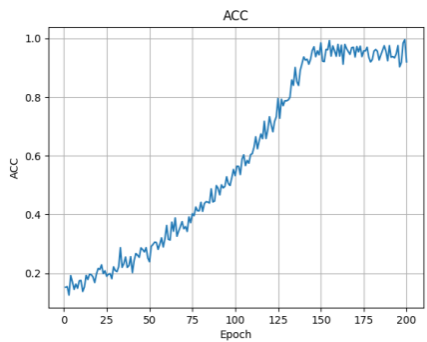

Figure 4 Accuracy curve during training

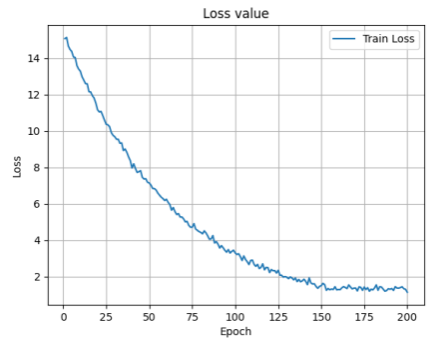

Figure 5 Loss function decline curve during training

As can be seen from Figures 4 and 5, our model reached convergence at approximately the 150th epoch, which means that after this training cycle, the performance indicators of the model, such as loss function values, accuracy, etc., no longer change significantly or Improvement, indicating that the model has learned most of the information available in the training data set. From this point, we can infer that during the first 150 epochs, the model went through a rapid learning phase, and its weights were continuously adjusted to minimize the loss function and improve the prediction accuracy of the training samples; and during the 150th epoch Around the time, the learning rate slows down significantly, and even if the training rounds continue to be increased, the performance of the model will not improve much. This is usually because the model begins to overfit the training data or has reached the optimal solution under the current architecture. Therefore, choosing to stop training at the 150th epoch can not only avoid the risk of overfitting, but also save computing resources and improve the efficiency of model training. In addition, this also reminds us that we can consider using early stopping strategy (Early Stopping) in future work, that is, terminating training early after the performance on the verification set stops improving, so as to further optimize the model training process and ensure that the model has good performance. Generalization ability. In summary, through the analysis of Figures 4 and 5, we not only confirmed the specific epoch number for model convergence, but also provided valuable direction guidance for subsequent model optimization.

## V. CONCLUSION

This study successfully improved the performance of YOLOv5s in the target detection task by introducing the knowledge distillation method. In particular, significant effect differences were observed under different distillation temperature conditions. Experiments show that as the distillation temperature increases, the mAP50 and mAP50-95 indicators of the student model increase, reaching a maximum of 96.75% and 74.56%, which are 5.42% and 6.7% higher than the YOLOv5s model without distillation respectively. In addition, the training curve of the model shows that the learning rate slows down significantly after about 150 epochs, indicating that the model has reached its optimal state. Therefore, reasonable setting of distillation temperature and timely adoption of early stopping strategies are crucial to avoid overfitting, save computing resources, and improve model training efficiency. The study's approach to integrating localization and classification knowledge, along with the innovative region-based selective distillation strategy, further enhances the robustness and adaptability of the model in diverse detection scenarios. Ultimately, the research provides a compelling framework for the advancement of object detection technologies, offering both theoretical insights and practical strategies for the deployment of more accurate and efficient deep learning models, while laying a solid groundwork for future exploration in optimizing target detection through knowledge distillation techniques.